\pdfoutput=1

\documentclass[11pt]{article}

\usepackage{coling}

\usepackage{times}
\usepackage{latexsym}

\usepackage[T1]{fontenc}

\usepackage[utf8]{inputenc}

\usepackage{microtype}

\usepackage{inconsolata}

\usepackage{graphicx}
\usepackage{booktabs}
\usepackage{multirow}
\usepackage{makecell}
\usepackage{pifont}
\usepackage{adjustbox}
\usepackage{rotating}
\usepackage{makecell}
\newcommand{\cmark}{\ding{51}} 

\title{Overview of the First Workshop on Language Models for \\ Low-Resource Languages (LoResLM 2025)}

\author{Hansi Hettiarachchi\textsuperscript{1}, Tharindu Ranasinghe\textsuperscript{1}, Paul Rayson\textsuperscript{1}, Ruslan Mitkov\textsuperscript{1}\\
{\bf Mohamed Gaber\textsuperscript{2}, Damith Premasiri\textsuperscript{1}, Fiona Anting Tan\textsuperscript{3}, Lasitha Uyangodage\textsuperscript{4}} \\ 
  \textsuperscript{1}Lancaster University, UK \textsuperscript{2}Birmingham City University, UK \\
  \textsuperscript{3}National University of Singapore, Singapore  \textsuperscript{4}University of Münster, Germany\\ 
  \texttt{loreslm2025@gmail.com}
 }

\begin{document}
\maketitle
\begin{abstract}


The first Workshop on Language Models for Low-Resource Languages (LoResLM 2025) was held in conjunction with the 31\textsuperscript{st} International Conference on Computational Linguistics (COLING 2025) in Abu Dhabi, United Arab Emirates. This workshop mainly aimed to provide a forum for researchers to share and discuss their ongoing work on language models (LMs) focusing on low-resource languages, following the recent advancements in neural language models and their linguistic biases towards high-resource languages. LoResLM 2025 attracted notable interest from the natural language processing (NLP) community, resulting in 35 accepted papers from 52 submissions. These contributions cover a broad range of low-resource languages from eight language families and 13 diverse research areas, paving the way for future possibilities and promoting linguistic inclusivity in NLP.


\end{abstract}

\section{Introduction}

Language models (LMs) have been a long-standing research topic, originating with simple n-gram models in the 1950s \cite{shannon1951prediction}. They are computational models that use the generative likelihood of word sequences to perform natural language processing (NLP) tasks \cite{zhao2023survey}. Recent advancements in LMs have significantly shifted towards neural language models due to their more robust capabilities \cite{zhao2023survey, minaee2024large}. Developing pre-trained neural language models/transformers is a key milestone in LM research that notably enhanced NLP performance \cite{vaswani2017attention, devlin2019bert}. This breakthrough has also prompted the development of more advanced large language models (LLMs), such as GPT, which consist of vast numbers of parameters pre-trained on extensive text corpora, resulting in state-of-the-art natural language understanding and generation across various applications \cite{touvron2023llama,jiang2023mistral}.

There are approximately 7,000 spoken languages worldwide \cite{van2022writing}. However, most NLP research focuses on about 20 languages with high resources \cite{magueresse2020low}. For example, 63\% of the papers published at ACL 2008 focused on English \cite{bender2011achieving}, and even a decade later, 70\% of the papers at ACL 2021 were evaluated only in English \cite{ruder2022square}. The remaining numerous languages that receive little research attention are commonly referred to as low-resource languages. These languages generally lack sufficient digital data and resources to support NLP tasks. They are also known as resource-scarce, resource-poor, less computerised, low-data, or low-density languages \cite{ranathunga2023neural}.

Since the capabilities of LMs are primarily determined by the characteristics of their pre-trained language corpora, disparities in language resources are also evident within the models. For instance, many widely used transformer models (e.g., \texttt{BERT} \cite{devlin2019bert}, \texttt{RoBERTa} \cite{liu2019roberta}, \texttt{BART} \cite{lewis-etal-2020-bart}, and \texttt{T5} \cite{10.5555/3455716.3455856}) only support English. However, the cross-lingual capabilities of transformers have paved the way for multilingual models (e.g., \texttt{mBERT} \cite{devlin2019bert}, \texttt{XLM-R} \cite{conneau2020unsupervised}, \texttt{mT5} \cite{xue-etal-2021-mt5}, and \texttt{BLOOM} \cite{workshop2022bloom}), allowing low-resource languages to benefit from other languages through joint learning approaches. Despite this progress, these models are typically limited to up to 100 languages due to the curse of multilingualism \cite{conneau2020unsupervised}. In light of this challenge, developing monolingual models (e.g., \texttt{SinBERT} for Sinhala \cite{dhananjaya-etal-2022-bertifying}, and \texttt{PhoBERT} for Vietnamese \cite{nguyen-tuan-nguyen-2020-phobert}) is another growing trend recently established to promote research in low-resource languages.

There are several common factors which impede low-resource language research. One major issue is limited data availability, as the performance of most models depends heavily on the amount of training data \cite{hettiarachchi-etal-2024-nsina}. Even recent neural LMs with multilingual capabilities tend to perform poorly when pre-training data for a particular language is limited or unseen \cite{ahuja2022multi, hettiarachchi2023ttl}. Data quality also plays a pivotal role in research outcomes, yet the absence of recommended guidelines hinders the quality of low-resource language data \cite{lignos2022toward}. Additionally, the scarcity of benchmark datasets tailored for low-resource languages tends to bias most model evaluations towards high-resource languages \cite{blasi2022systematic, ranasinghe2022sold}.


Interestingly, there are several ongoing efforts that aim to encourage research on low-resource languages and mitigate the bias in NLP approaches towards high-resource languages \cite{ltedi2022language, loresmt2023technologies, sigul2024special}. We organised the first Workshop on Language Models for Low-Resource Languages (LoResLM 2025) to further strengthen this trend. LoResLM 2025\footnote{Available at \url{https://loreslm.github.io/}} specifically focused on LM-based approaches for low-resource languages, inviting submissions on a broad range of topics, including creating corpora, developing benchmarks, building or adapting LMs, and exploring LM applications for low-resource languages. Section \ref{sec:workshop-contr} provides a summary of the workshop contributions, highlighting language and task/research area coverage. We invite you to refer to the full papers available in the proceedings for more detailed information.



\section{Workshop Contributions} \label{sec:workshop-contr}


LoResLM 2025 received 52 submissions, including 40 long papers and 12 short papers. Among these, we accepted 35 papers, including 28 long papers and seven short papers, to appear in the workshop proceedings, following the review process. We provide a detailed summary of the distribution of accepted papers across various languages and research areas below.

\subsection{Languages}

\begin{figure}[!hbt]
\centering
\includegraphics[width=\columnwidth]{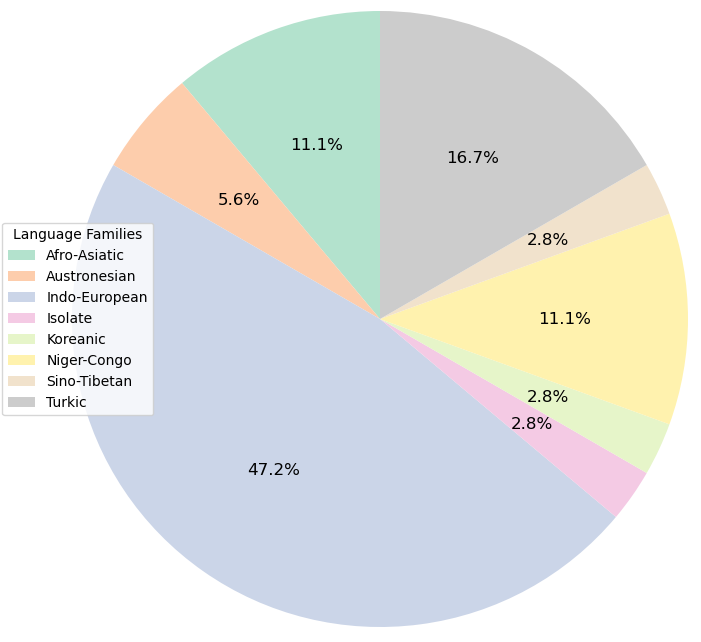}
\caption{Distribution of workshop contributions across language families}
\label{fig:lang-pie}
\end{figure}

As illustrated in Figure \ref{fig:lang-pie}, the papers accepted to LoResLM 2025 mainly span eight language families. The majority representation is from Indo-European family, while Koreanic, Sino-Tibetan and Isolate language families have equal minority representation. Languages with no relationships with others were considered under the Isolate family.

We present a detailed language-level analysis in Table \ref{tab:lang-coverage}. We further divided the Indo-European family into its first branch level for a comprehensive exploration, given its wide contributions. Overall, there were contributions from four distinct branches of the Indo-European language family. During this analysis, we focused exclusively on low-resource languages, excluding high-resource languages involved in comparison studies. However, some languages that would typically classify as high-resource considering the general resource distribution across popular research areas (e.g. Arabic, German, etc.) were considered low-resource in specific contexts where resources are limited, such as particular domains, research areas, or dialects. In total, contributions covered 28 low-resource languages. Additionally, a few papers experimented with multiple languages (more than five) from various language families. These were categorised under `Multiple' but excluded from the language count given above, as their focus was more on the task level rather than the language level.  

\renewcommand{\arraystretch}{1}
\begin{table*}[!h]
  \centering
  \scalebox{0.92}{
  \begin{tabular}{llp{0.68\linewidth}}
    \toprule
    \bf Language Family & \bf Language & \bf Papers \\
    \hline
    \multirow{2}{*}{Afro-Asiatic} & Arabic & \citet{nacar-etal-2025-arabic, shang-etal-2025-atlas-chat, zeinalipour-etal-2025-arabic-crosswords}\\
    & Hausa & \citet{sani-etal-2025-hausa}\\
    \hline
    \multirow{2}{*}{Austronesian} & Filipino & \citet{gamboa-lee-2025-filipino}\\
    & Tagalog & \citet{cruz-2025-transformers}\\
    \hline
    \multirow{2}{*}{\makecell[l]{Indo-European\\(Germanic)}} & German & \citet{zhukova-etal-2025-semantic-search}\\
    & Old English & \citet{harju-etal-2025-age-bert}\\
    \hline
    \makecell[l]{Indo-European\\(Hellenic)} & Ancient Greek & \citet{rapacz-smywinski-pohl-2025-interlinear}\\
    \hline
    \multirow{5}{*}{\makecell[l]{Indo-European\\(Indo-Iranian)}} & Bengali & \citet{alam-etal-2025-bnsentmix, sadhu-etal-2025-bangla-bias}\\
    & Marathi & \citet{mutsaddi-choudhary-2025-plagiarism, dmonte-etal-2025-mt-offensive}\\
    & Persian & \citet{habibzadeh-asadpour-2025-persian, mokhtarabadi-etal-2025-persian, zeinalipour-etal-2025-persianmcq} \\
    & Sinhala & \citet{dmonte-etal-2025-mt-offensive}\\
    & Urdu & \citet{amin-etal-2025-drs, donthi-etal-2025-idiomatic}\\
    \hline
    \multirow{4}{*}{\makecell[l]{Indo-European\\(Italic)}} & Italian & \citet{amin-etal-2025-drs}\\
    & Medieval Latin & \citet{liu-etal-2025-comparative}\\
    & Monégasque & \citet{merad-etal-2025-language}\\
    & Portuguese & \citet{lasheras-pinheiro-2025-calquest}\\
    \hline
    Isolate & Basque & \citet{kryvosheieva-levy-2025-syntactic}\\
    \hline
    Koreanic & Korean & \citet{tran-etal-2025-wsd}\\
    \hline
    \multirow{4}{*}{Niger-Congo} & isiXhosa & \citet{matzopoulos-etal-2025-babylms}\\
    & IsiZulu & \citet{mahlaza-etal-2025-isizulu}\\
    & Mooré & \citet{ouattara-etal-2025-literacy}\\
    & Swahili & \citet{kryvosheieva-levy-2025-syntactic}\\
    \hline
    Sino-Tibetan & Cantonese & \citet{dai-etal-2025-cantonese}\\
    \hline
    \multirow{4}{*}{Turkic} & Kazakh & \citet{veitsman-hartmann-2025-turkic}\\
    & Kyrgyz & \citet{veitsman-hartmann-2025-turkic}\\
    & Turkish & \citet{veitsman-hartmann-2025-turkic}\\
    & Turkmen & \citet{veitsman-hartmann-2025-turkic}\\
    & Uzbek & \citet{veitsman-hartmann-2025-turkic, bobojonova-etal-2025-bbpos}\\
    \hline
    Multiple &  & \citet{bagheri-etal-2025-beyond, zhu-etal-2025-incontext, tashu-tudor-2025-crosslingual, sindhujan-etal-2025-translation, dewangan-etal-2025-segmentation}\\
    \bottomrule
  \end{tabular}
  }
  \caption{
    Coverage of workshop papers across different languages. The final row (`Multiple') represents the scenario where more than five languages from multiple language families are experimented with.
  }
  \label{tab:lang-coverage}
\end{table*}

\subsection{Research Areas}

Table \ref{tab:nlp_contributions} shows the distribution of the accepted papers across various NLP research areas. These areas were adopted based on the topics of call for papers from leading NLP conferences in 2024. 

Overall, the accepted papers contributed to 13 NLP research areas. As expected, the most popular topic among the accepted papers was \textit{`Language Modelling'} with eleven papers.  \textit{`Machine Translation and Translation Aids'} was the second most popular topic with six papers. The other topics approximately had a similar number of papers. Apart from the papers mentioned in  Table \ref{tab:nlp_contributions}, \citet{veitsman-hartmann-2025-turkic} provided a survey on Central Asian Turkic languages spanning across several research areas. 

\begin{table*}[h!]
\centering
\small
\resizebox{\textwidth}{!}{
\begin{tabular}{l|c|c|c|c|c|c|c|c|c|c|c|c|c|}
\toprule
\multicolumn{1}{c|}{Paper} & \rotatebox{90}{\makecell[l]{Dialogue and \\ Interactive Systems}} & \rotatebox{90}{\makecell[l]{Ethics, Bias, \\ and Fairness}} & \rotatebox{90}{\makecell[l]{Information Retrieval \\ and Text Mining}} & \rotatebox{90}{Language Modelling} & \rotatebox{90}{\makecell[l]{Linguistic Insights Derived \\ using Computational Techniques}} & \rotatebox{90}{\makecell[l]{Machine Translation and \\ Translation Aids}} & \rotatebox{90}{NLP and LLM Applications} & \rotatebox{90}{\makecell[l]{Offensive Speech Detection \\ and Analysis}} & \rotatebox{90}{\makecell[l]{Phonology, Morphology and \\ Word Segmentation}} & \rotatebox{90}{Question Answering} & \rotatebox{90}{Lexical Semantics} & \rotatebox{90}{\makecell[l]{Sentiment Analysis, Stylistic Analysis, \\ Opinion and Argument Mining}} & \rotatebox{90}{\makecell[l]{Syntactic analysis \\ (Tagging, Chunking, Parsing)}} \\

\midrule
\citet{liu-etal-2025-comparative} & & & & & & & & & & & \cmark & & \\
\citet{gamboa-lee-2025-filipino} &  & \cmark & & & & & & & & & & & \\
\citet{alam-etal-2025-bnsentmix} & & & & & & & & & & & & \cmark & \\
\citet{cruz-2025-transformers} & &  & & \cmark & & & & & & & & & \\
\citet{dai-etal-2025-cantonese} & & & & & & \cmark & & & & & & & \\
\citet{turumtaev-2025-adaptive} & &  & & \cmark & & & & & & & & & \\
\citet{sani-etal-2025-hausa} & & & & & & & & & & & \cmark & & \\
\citet{mutsaddi-choudhary-2025-plagiarism} & & & &  & & & \cmark & & & & & & \\
\citet{amin-etal-2025-drs} & & & & & & & & &  & & & & \cmark \\
\citet{bagheri-etal-2025-beyond} & & & & \cmark & & & & & & & & & \\
\citet{ouattara-etal-2025-literacy} & \cmark & & & & & & & & & & & & \\
\citet{zhu-etal-2025-incontext} & & &  & & \cmark & & & & & & & & \\
\citet{matzopoulos-etal-2025-babylms} & &  & & \cmark & & & & & & & & & \\
\citet{rapacz-smywinski-pohl-2025-interlinear} & & &  & & & \cmark & & & & & & & \\
\citet{habibzadeh-asadpour-2025-persian} & & & & & & & &  & & & & \cmark & \\
\citet{dmonte-etal-2025-mt-offensive} & & &  & &  & \cmark & & \cmark & & & & & \\
\citet{tashu-tudor-2025-crosslingual} & &  & & \cmark & & & & & & & & & \\
\citet{mokhtarabadi-etal-2025-persian} & &  & & \cmark & & & & & & & & & \\
\citet{tran-etal-2025-wsd} & & & & & & & & & & & \cmark  & & \\
\citet{merad-etal-2025-language} & & &  & & & \cmark & & & & & & & \\
\citet{mahlaza-etal-2025-isizulu} & & &  & & \cmark & & & & & & & & \\
\citet{nacar-etal-2025-arabic} & &  & & \cmark & & & & & & & & & \\
\citet{kryvosheieva-levy-2025-syntactic} & &  & & \cmark &  & & & & & & & & \\
\citet{harju-etal-2025-age-bert} & &  & & \cmark & & & & & & & & & \\
\citet{shang-etal-2025-atlas-chat} & &  & & \cmark & & & & & & & & & \\
\citet{donthi-etal-2025-idiomatic} & & &  &  & & \cmark & & & & & & & \\
\citet{sadhu-etal-2025-bangla-bias} &  & \cmark & &  & & & & & & & & & \\
\citet{sindhujan-etal-2025-translation} & & &  & & & \cmark & & & & & & & \\
\citet{bobojonova-etal-2025-bbpos} & & & & & & & & & & & & & \cmark  \\
\citet{dewangan-etal-2025-segmentation} & & & & & & & &  & \cmark & & & & \\
\citet{zeinalipour-etal-2025-persianmcq} & & & & &  & & & & & \cmark & & & \\
\citet{lasheras-pinheiro-2025-calquest} & & & & &  & & & & & \cmark & & & \\
\citet{zeinalipour-etal-2025-arabic-crosswords} & & & &  & & & \cmark & & & & & & \\
\citet{zhukova-etal-2025-semantic-search} & & & \cmark  & & & & & & & & & & \\

\bottomrule
\end{tabular}
}
\caption{Coverage of workshop papers across different NLP areas.}
\label{tab:nlp_contributions}
\end{table*}

\section{Conclusions}
The first Workshop on Language Models for Low-Resource Languages (LoResLM 2025) attracted a lot of interest from the NLP community, having 35 accepted papers from 52 submissions. The accepted papers mainly span eight language families, with the majority representation being from Indo-European families. Furthermore, the accepted papers contributed to 13 NLP research areas, with major contributions to \textit{`Language Modelling'}  and \textit{`Machine Translation and Translation Aids'}. We believe the findings and resources from LoResLM will open exciting new avenues to empower linguistic diversity for millions of low-resource languages. 

For the future iterations of LoResLM, we expect better representation from more diverse linguistic groups, particularly those from underrepresented families such as Uralic, Dravidian and Indigenous languages of the Americas. Furthermore, we aim to diversify research topics, encouraging work in areas such as speech processing, information extraction, and dialogue systems, which are critical for many practical applications.

\bibliography{custom}

\end{document}